\documentclass[conference]{IEEEtran}
\IEEEoverridecommandlockouts
\usepackage{cite}
\usepackage{amsmath,amssymb,amsfonts}
\usepackage{algorithmic}
\usepackage{graphicx}
\usepackage{mathtools}
\usepackage{amsmath,amssymb}
\newlength\mylength
\usepackage{hyperref}
\usepackage{xcolor}
\usepackage{textcomp}
\usepackage{tabularx}
\usepackage{siunitx}
\usepackage{booktabs}
\usepackage{caption}
\usepackage{subcaption}
\usepackage[utf8]{inputenc}
\usepackage{array}
\usepackage{collcell}
\usepackage{datatool}
\usepackage{algorithm}
\usepackage{algorithmic}

\DeclareCaptionFont{9pt}{\fontsize{9pt}{10pt}\selectfont}
\usepackage{xcolor}
\usepackage{booktabs}
\usepackage{lipsum}
\usepackage{mathtools}
\usepackage{amsmath}
\usepackage{mathtools}
\usepackage[justification=centering]{caption}
\usepackage[mathscr]{euscript}
\usepackage{cuted}
\DeclareMathOperator{\EX}{\mathbb{E}}
\def\BibTeX{{\rm B\kern-.05em{\sc i\kern-.025em b}\kern-.08em
    T\kern-.1667em\lower.7ex\hbox{E}\kern-.125emX}}
    
\begin{document}

\title{Missing Data Estimation in Temporal Multilayer Position-aware Graph Neural Network (TMP-GNN) \\
{}
}

\author{\IEEEauthorblockN{ Bahareh Najafi\ \  }
\IEEEauthorblockA{\textit{University of Toronto}\\
Toronto, Canada \\
bahareh.najafi@mail.utoronto.ca}
\and
\IEEEauthorblockN{Saeedeh Parsaeefard}
\IEEEauthorblockA{\textit{University of Toronto}\\
Toronto, Canada \\
saeideh.fard@utoronto.ca}
\and
\IEEEauthorblockN{Alberto Leon-Garcia}
\IEEEauthorblockA{\textit{University of Toronto}\\
Toronto, Canada \\
alberto.leongarcia@utoronto.ca}
}\maketitle

\begin{abstract}
GNNs have been proven to perform highly effective in various node-level, edge-level, and graph-level prediction tasks in several domains. Existing approaches mainly focus on static graphs. However, many graphs change over time with their edge may disappear, or node$/$edge attribute may alter from one time to the other. It is essential to consider such evolution in representation learning of nodes in time varying graphs. In this paper, we propose a Temporal Multi-layered Position-aware Graph Neural Network (TMP-GNN), a node embedding approach for dynamic graph that incorporates the interdependence of temporal relations into embedding computation. We evaluate the performance of TMP-GNN on two different representations of temporal multilayered graphs. The performance is assessed against most popular GNNs on node-level prediction task. Then, we incorporate TMP-GNN into a deep learning framework to estimate missing data and compare the performance with their corresponding competent GNNs from our former experiment, and a baseline method. Experimental results on four real-world datasets yields up to $58\%$ of lower $\text{ROC AUC}$ for pairwise node classification task, and $96\%$ of lower $\text{MAE}$ in missing feature estimation, particularly for graphs with relatively high number of nodes and lower mean degree of connectivity.
\end{abstract}

\begin{IEEEkeywords}
Missing Data Analysis, Node Embedding, Position-aware Graph Neural Network, Spatio-temporal Measurements, Temporal Multi-Layer Graph
\end{IEEEkeywords}

\section{Introduction}
Graph neural networks (GNN) has recently been used as a standard in developing machine learning methods for graphs. Graphs that have been formed from transportation network \cite{lv2014traffic},\cite{lin2018pattern},\cite{cui2019traffic}, \cite{yu2017spatiotemporal}, brain network \cite{ghoroghchian2020node}, social media community networks, etc. The GNN architecture has effectively combined the node$/$edge features and graph topology to build distributed representation. The resulting representation can be used to solve node-level, edge-level \cite{zhang2018link} and graph-level prediction tasks \cite{ghoroghchian2020node}. 
 \par The goal of node embedding methods is to identify a vector representation that captures node location within a broader topological structure of the graph. Most node embeddings learned from GNN architectures focus on single static graphs. These methods assume that the number/position of nodes as well as their interactions do not change over time.
However, in many applications we tackle with spatio-temporal measurements collecting over a time span wherein the necessity of time varying graphs emerge. In such graphs, the number of nodes, their connecting edges, and the edge weights vary from time to time, thus dynamic node embedding should be learned accordingly from the graph. \par Time varying graphs can be defined as continuous or discrete time dynamic graphs which the latter can be represented as a sequence of interdependent time layers; Each layer is a graph building from existing nodes and weighted edges corresponding to a given time. 
One straightforward way of computing node embedding in discrete time varying graphs is to use a static GNN based node embedding for each individual time layer and aggregate the results of corresponding layers through recurrent neural network variants. However, this method would compute an embedding for each layer independently, and consequently ignore the inter-layer correlation.  In this paper we present TMP-GNN, a temporal multilayer position-aware GNN based node embedding which is an extension of its static version position-aware GNN (P-GNN) \cite{you2019position}.
The goal of P-GNN is to learn position-aware node embedding that utilize local network structure and the global network position of a given node with respect to randomly selected nodes called anchor-sets that enables us to distinguish among isomorphic nodes; nodes that are based in far different parts of the graph but have topologically same structure.  The distinguishing power will improve if the node/edge features are available. The resulted embeddings can be later used to approximately calculate the shortest path distance among the embedded nodes in the graph.
The contributions of the paper are summarized as follows: \par

(1)	We learn the short-term temporal dependencies, global position, and feature information of the graph jointly through our TMP-GNN embedding component and utilize the derived representation in missing data estimation framework.
\par (2) Instead of modeling a dynamic graph via multi-graph, we exploit a supra-adjacency matrix to encode a temporal graph with its intra-layer and inter-layer coupling in a single graph that facilitate faster learning and higher Area Under the Receiver Operating Characteristic (ROC) Curve (AUC) for pairwise node classification.
\par(3)	We deploy the concept of conditional centrality derived from eigenvector-based centrality to distinguish nodes of higher influence and integrate it in message aggregation across the graph.
\par(4)	We identify a new experimental setting to enhance the training of multi-graph deployment of dynamic graph so that it can better trace the behavioural changes of nodes in the graph.
\par (5) We use hidden states learned from bi-directional GRU (bi-GRU) to learn the long-term temporal dependencies in both forward and backward directions to estimate missing values.
\par(6) We conduct several experiments using four real-world datasets, with a wide range of node number, degree of connectivity, edge dynamic and area of study.  The results illustrate that TMP-GNN improve the $\text{ROC AUC}$ significantly as compared to the best baseline, and reduce the MAE of missing data estimation. 

\section{Notations and Preliminaries} \label{pre}
Before diving deeply into details of our proposed architecture, we provide some backgrounds on temporal multi-layer graphs. 
\subsection{Notation} 
Fig. \ref{fig:fig114} illustrates a temporal multi-layer graph. We can represent the graph as $\mathcal{G} = (\mathcal{V}, {{\rho}^{(\mathcal{E} \times t)},\mathcal{E}^{(t)}},\tilde{\mathscr{E}})$, where $\mathcal{V}$ are the set of nodes $\{v,\forall v \in \mathcal{V}\}$. 
Here, ${\rho}^{(\mathcal{E} \times t)}$ indicate whether a given edge is present at a given time $t$. $\mathcal{E}^{(t)}$ shows intra-layer weighted edges at time $t$, $\mathcal{E}^{(t)} = \{e_{uv}^{(t)}, \forall u,v \in \mathcal{V}\}$. $\tilde{\mathscr{E}}$ indicates inter-layer edges $\{(t,t',\tilde{x}_{tt'})\in \tilde{\varepsilon}\}$, where $\tilde{x}_{tt'}$ is the edge weight between time layers $(t,t').$ The multi-layer graph is also denoted by a sequence of adjacency matrices $\boldsymbol{ {A^{(t)}} } \in \mathbb{R}^{\mathrm{N} \times \mathrm{N}}$, where $A_{uv}^{(t)}$ denotes directed edge from node $u$ to $v$ in time layer $t$, as follows \label{eq1}:

\begin{figure*}[!t]
\centering
\includegraphics[scale=0.85]{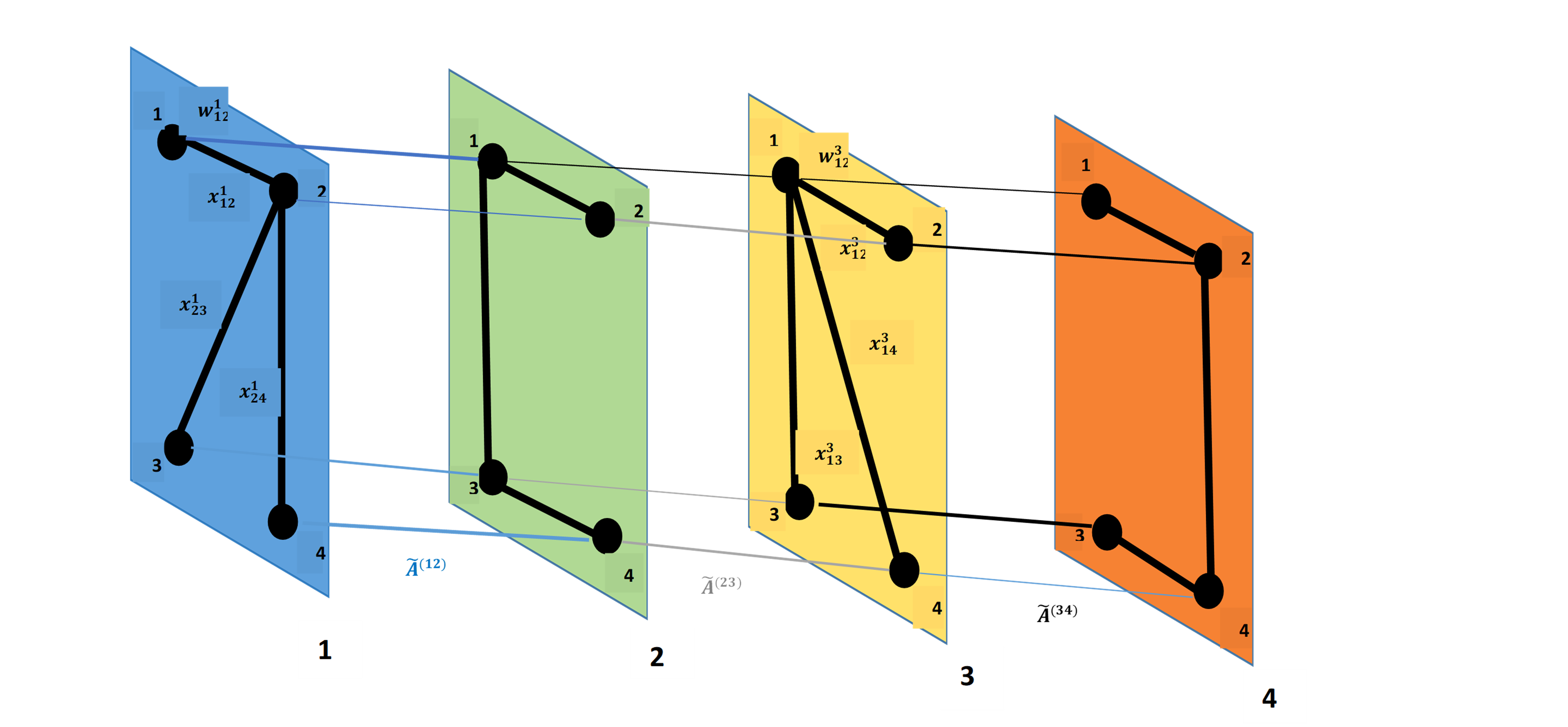}
\caption{A dynamic multi-layer graph}
\label{fig:fig114}
\end{figure*}

\begin{equation}
\label{eq1}
A_{uv}^{(t)} = \begin{cases}
              x_{uv}^{(t)},  & \textnormal{if} \ \  \textnormal{$u,v$ are connected at time $t$} \\
             0,  & \textnormal{otherwise.}
       \end{cases} 
\end{equation}
where $x_{uv}^{(t)}$ is the feature vector associated to $e_{uv}^{(t)}$.  In scenarios where the node attributes are available, another set is defined as $ W^{(t)} = \{ w_v^{(t)}, \forall t \in \mathbb{N}_{T}\}$ where $w_v^{(t)}$ is the feature vector associated to node $v$ in time layer $t$. \par The set $\tilde{\mathscr{E}}$ encodes the weight, and topology of the coupling of individual instances of the same nodes between pairs of time layers ${(t, t')}$. 


The inter-layer coupling described above is diagonal and uniform \cite{taylor2017eigenvector} which means the existence of inter-layer edge is restricted between separate instances of same nodes from one layer to another. For instance, there is no edge from node $v$ in $t$ to node $u$ in $t'$. The corresponding layers are uniformly coupled, meaning that the inter-layer edge weight between two layers is identical for all nodes in those layers. Please note that this assumption can be generalized so that different inter-layer edge weights are assigned to different subsets of nodes; that is $\tilde{x}_{tt'}^{u} \neq \tilde{x}_{tt'}^{v}$. \par There are two possible ways to form the coupling between the layers: directed and undirected chain; The former couples the adjacent time layers by neglecting the directionality of time as following equation:
\begin{equation}
\label{eq3.1}
\tilde{A}_{uv}^{(tt')} = \left\{
     \begin{array}{@{}l@{\thinspace}l}
       1  &: \forall $u,v$  \textnormal{if} \ \ $u=v$,  |$t'-t$|=\delta \\
       0 &:  \textnormal{otherwise} \\
     \end{array}
   \right.
\end{equation} 

\noindent where $\delta$ is the sampling rate. The latter not only respects the time direction, but also include weighted undirected connection between same nodes of all pair of time layers as in (\ref{eq3.2}).
\begin{equation}
\label{eq3.2}
\tilde{A}_{uv}^{(tt')}=\left\{\begin{array}{ll}
                1+\gamma, \forall $u,v$ & \textnormal{if} \ \, $u=v$, |$t'-t$|=\delta \\
                \gamma  & \textnormal{otherwise},
            \end{array}\right.
\end{equation} 
 
 \noindent where ${\boldsymbol{\tilde{A}}}$ is a $\mathrm{T} \times \mathrm{T}$ inter-layer matrix between, and $\gamma$ is node teleportation probability \cite{gleich2015pagerank}. In this paper, we have used the former approach for inter-layer coupling. The inter-layer coupling of same nodes decrease as $|t'-t|$ increases, and we adjust $\delta$ according to data characteristics and inter-sequence correlation in our experiments.

A node embedding function e.g., SAGE, can be represented as $ f  \colon \mathcal{V}^{(t)}  \rightarrow \mathcal{Z}^{(t)}$  that maps a given node $v$ to d-dimensional vector $\mathbf{z}$ in layer $t$ where $\mathcal{Z}^{(t)} = \{\mathbf{z}_v^{(t)}, \forall v,t \in \mathbb{N}_{|\mathcal{V}|}, \mathbb{N}_T \},\mathbf{z}_v^{(t)} \in \mathbb{R}^d$. Once the node embedding is computed, the corresponding edge embedding can be calculated as $\mathbf{z}_{uv}^{(t)} = {g}(\mathbf{z}_u^{(t)}, \mathbf{z}_v^{(t)})$, where $g$ function is equal to $\text{mean}$ in our case of study.

\subsection{Supracentrality Matrix for Temporal Multilayer Position-aware Graph } \label{supra}

As explained, a temporal graph is represented through a sequence of adjacency matrices that each of which refers to one layer of a dynamic network at a specific point of time. Then, we construct a supracentrality matrix $\boldsymbol{\mathbb{C}}(\omega)$ by linking up centrality matrices across time layers through a weighted inter-layer parameter called $\omega$ that is used to adjust the extent of coupling strength among pair of time layers. The entries of dominant eigenvector of $\mathbb{C}(\omega)$ shows joint centrality, the importance of every node-layer pair ${(v, t)}$. Additionally marginal and conditional centralities are defined to represent the relative importance of a node compared to other nodes at time layer $t$. \par  The supracentrality framework is mainly focused on eigenvector-based centrality, that is obtained by calculating the centralities as the elements of the dominant eigenvector corresponding to the largest-magnitude eigenvalue of a centrality matrix $\mathcal{C}(\boldsymbol{A})$ that is defined as a function of network adjacency matrix $\boldsymbol{A}$. We have selected centrality measure from one of the most popular choices to be equal to adjacency matrix ($\mathcal{C}(\boldsymbol{A}) = \boldsymbol{A}$).  

\par $\mathbb{C}(\omega)$, a supracentrality matrix for a dynamic graph is a group of matrices formed as below:

\begin{equation}
\label{eq4}
\mathbb{C}(\omega)=\hat{\mathbb{C}}+\omega\hat{\mathbb{A}}, 
\end{equation}
where
\begin{eqnarray}
\hat{\mathbb{C}} = \begin{bmatrix}
\boldsymbol{C}^{(1)} & 0 & 0 & \cdots & 0\\
0 & \boldsymbol{C}^{(2)} & 0 & \cdots & 0\\
0 & 0 & \boldsymbol{C}^{(3)} & \cdots & 0\\
\cdots & \cdots & \cdots & \cdots &\cdots \\
0 & 0 & 0 & \cdots &  \boldsymbol{C}^{(T)}
\end{bmatrix}, \nonumber 
\end{eqnarray}
and
\begin{eqnarray}
\hat{\mathbb{A}} =
\begin{bmatrix}
\tilde{A}^{(11)}\boldsymbol{I} & \tilde{A}^{(12)}\boldsymbol{I} & \tilde{A}^{(13)}\boldsymbol{I} & \cdots & \tilde{A}^{(1T)}\boldsymbol{I}\\
\tilde{A}^{(21)}\boldsymbol{I} & \tilde{A}^{(22)}\boldsymbol{I} & \tilde{A}^{(23)}\boldsymbol{I} & \cdots & \tilde{A}^{(2T)}\boldsymbol{I}\\
\tilde{A}^{(31)}\boldsymbol{I} & \tilde{A}^{(32)}\boldsymbol{I} & \tilde{A}^{(33)}\boldsymbol{I} & \cdots & \tilde{A}^{(3T)}\boldsymbol{I}\\
\cdots & \cdots & \cdots & \cdots &\cdots \\
\tilde{A}^{(T1)}\boldsymbol{I} & \tilde{A}^{(T2)}\boldsymbol{I} & \tilde{A}^{(T3)}\boldsymbol{I} & \cdots &  \tilde{A}^{(TT)}\boldsymbol{I}
\end{bmatrix}. \nonumber 
\end{eqnarray}  
$$$$

\noindent Here, $\mathbb{C}(\omega)$ consists of two components in (\ref{eq4}): $\hat{\mathbb{C}}$ and $\omega\hat{\mathbb{A}}$. The former  (diagonal)  component, $Diag[\mathbf{C}^{(1)},\cdots, \mathbf{C}^{(T)}]$, represents 
a set of $\mathrm{T}$ weighted centrality matrix 
of individual layers, and the latter (off-diagonal block), $\hat{\mathbb{A}}={\boldsymbol{\tilde{A}}}\otimes \boldsymbol{I}$ encodes the uniform and diagonal coupling with strength parameter $\omega$ between the time layers, where  ${\boldsymbol{\tilde{A}}}$ is defined in (\ref{eq3.1}), $\otimes$ is the kronicker product, and $\boldsymbol{I}$ is a $\mathrm{N} \times \mathrm{N}$ identity matrix, since we only consider the coupling among same nodes between consecutive pair of layers. We study the dominant eigenvector $\mathbb{V}(\omega)$ of $\mathbb{C}(\omega)$ corresponding to largest eigenvalue $\lambda_{\max}$ as in \ref{eq9}. 

\begin{equation}
\label{eq9}
\mathbb{C}(\omega)\mathbb{V}(\omega)=\lambda_{\max}(\omega)\mathbb{V}(\omega).
\end{equation}

\noindent The elements in the dominant right eigenvector of $ \mathbb{V}(\omega)$ are interpreted as scores that measure the importance of node-layer pairs ${(v, t)}$. As such, the eigenvalue entity ${\mathrm{W}}_{v,t}(\omega)$ in the following equation (\ref{eq10}) called joint centrality is a d-dimensional vector, reflecting the centrality of node $v$ at layer $t$, where $\mathbb{V}_{{\mathrm{N}}(t-1)+n}(\omega)$ is the ${{\mathrm{N}}(t-1)+n}$-{th} entity of largest eigenvalue $\mathbb{V}(\omega)$ and $n$ refers to node order $v_n$ that is removed here for simplicity. 

\begin{equation}
\label{eq10}
\mathrm{W}_{v,t}{(\omega)}=\mathbb{V}_{\textrm{N}(t-1)+n}{(\omega)}.
\end{equation}

Inspired by probability theory, the authors in \cite{taylor2017eigenvector} defined Marginal Layer Centrality $\mathrm{(MLC)}$ and Conditional centrality in following equations (\ref{eq11}, \ref{eq12}). $\mathrm{MLC}$ indicates the average joint centralities over all nodes at time layer $t$. Conditional centrality of node $v$ shows the importance of the node relative to other nodes at layer $t$.

\begin{equation}
\label{eq11}
\mathrm{MLC}_{v}(\omega)=\sum_{v = 1}^{{|\mathcal{V}|}}\mathrm{W}_{v,t}(\omega),
\end{equation}

\begin{equation}
\label{eq12}
CC_{v}(\omega)=\frac{\mathrm{W}_{v,t}(\omega)}{\mathrm{MLC}_{v}(\omega)}.
\end{equation}


\subsubsection{Choice of $\omega$ for Steady State Node Ranking} \label{omega}

\par A number of factors should be taken into account while choosing the value of $\omega$. The choice of $\omega$ should be appropriate for the targeted application or the research question.
When $\omega$ $\rightarrow \infty$, the centrality measures change slowly over time. On the other hand, in presence of small $\omega$, the $CC_{v}(\omega)$ fluctuates from one time layer to the other. 
We considered $\omega$ at strong coupling regime. If $\omega$ is too large (e.g. $\omega$ $>$ 100), the $CC_{v}(\omega)$ reaches to stationary that is equal to average $CC_{v}(\omega)$ over all time layers \cite{shai2017case}.
Moreover, the dominant eigenvalue of $\mathbb{C}(\omega)$ converge to the same values of $\boldsymbol{\hat{A}}$. That is, $\lambda_{\max}(\omega)\rightarrow \tilde{\mu}_1$, where $\lambda_{\max}$ and $\tilde{\mu}_1$ are the dominant eigenvalues of $\mathbb{C}(\omega)$ and $\boldsymbol{\hat{A}}$, respectively.

\begin{algorithm}
\caption{Find the appropriate value of $\omega$}\label{alg_1}
\begin{algorithmic} 
\REQUIRE Initialize the value of $\omega$ for {$\omega \geq 0$} {(we start from $\omega=10$)}, construct $\boldsymbol{B} = \omega\hat{\mathbb{A}}$, $\boldsymbol{C} = \mathbb{C}(\omega)$ \newline
 choose a random number for $b_k$, $c_{k'}$ as the initial values for the largest eigenvector of $\boldsymbol{B}$ and $\boldsymbol{C}$. We set $b_k$, $c_{k'}$ = $[1 1 \cdots 1]$ with its length equal to number of columns of $\boldsymbol{B}$, $\boldsymbol{C}$ which is $N \times T$ in our case.

\STATE $b_{k} \leftarrow \frac{b_k}{||b_k||}$, $c_{k'} \leftarrow \frac{c_{k'}}{||c_{k'}||}$
   \STATE $\lambda_{k} \leftarrow {b^{*}_k}{\boldsymbol{B}}{b_k}$, $\mu_{k'} \leftarrow {c^{*}_k}{\boldsymbol{C}}{c_k}$
   \STATE   k=0
\WHILE{$|\lambda_{k} - \mu_{k'}| < 0.01$}
\WHILE{$|\lambda_{k + 1} - \lambda_{k}| < 0.01$}
   \STATE $k \leftarrow k+1$ \newline
   \STATE  $b_{k}=\frac{\boldsymbol{B}{b_k}}{||\boldsymbol{B}{b_k}||}$ \newline
  \STATE $\lambda_{k}={b^{*}_{k}}{\boldsymbol{B}}{b_{k}}$
\ENDWHILE
\RETURN $\lambda_{k}$, ${b_{k}}$
\WHILE{$|\mu_{{k'} + 1} - \mu_{k'}| < 0.01$}
   \STATE $k' \leftarrow {k'}+1$ \newline
   \STATE $c_{k'}=\frac{\boldsymbol{C}{c_{k'}}}{||\boldsymbol{C}{c_{k'}}||}$ \newline
  \STATE $\mu_{k'}={c^{*}_{k'}}{\boldsymbol{C}}{c_{k'}}$
\ENDWHILE
\RETURN $\mu_{k'}$, ${c_{k'}}$
\ENDWHILE
\RETURN $\omega$
\end{algorithmic}
\end{algorithm}

\par Through power iteration method of eigenvector measurements \cite{ipsen2005analysis}, we found minimum value of $\omega$ that satisfies the above requirements (algorithm \ref{alg_1}). In the next section, we utilize stationary ${CC}_{v}(\omega)$ to distinguish highly important nodes belonging to P-GNN anchor-sets, and then exploit it to aggregate the information across the anchor-sets.

\subsection{TMP-GNN: Multi-layer Position-aware based Graph Neural Network} \label{PGNN}

Several GNN methods are distinguished in the way they aggregate nodes’ information from their neighborhood to form node representation at each layer k.  The approach can be summarized in two functions: AGGREGATE and COMBINE \cite{xu2018powerful}, which the former decides on how the information is aggregated from adjacent nodes, and the later updates the node representation from the previous layer $k-1$ to layer $k$. The underlying functions can be combined and jointly represented for some GNN, as follows:\newline 
\begin{flalign}
\label{eq15}
\mathbf{a}_{v_n}^{(k)}  &=  \textrm{AGGREGATE}^{(k)} (\lbrace  
\mathbf{h}_{u_n}^{(k-1)}:u_n\in \text{neighbours}(v_n) \rbrace),  \nonumber \newline\\ 
\mathbf{h}_{v_n}^{(k)} &=  \textrm{COMBINE}^{(k)}(\mathbf{h}_{v_n}^{(k-1)},\mathbf{a}_{v_n}^{(k)}),
\end{flalign}
where $h_{v}^{(k)}$ is the feature representation vector of node $v$ at iteration k/k-th layer of GNN, and $\text{Neighbours}(v)$  is the set of nodes adjacent to $v$  in $\mathcal{G}$.

\begin{figure*}[!t]
\centering
\includegraphics[scale=0.18]{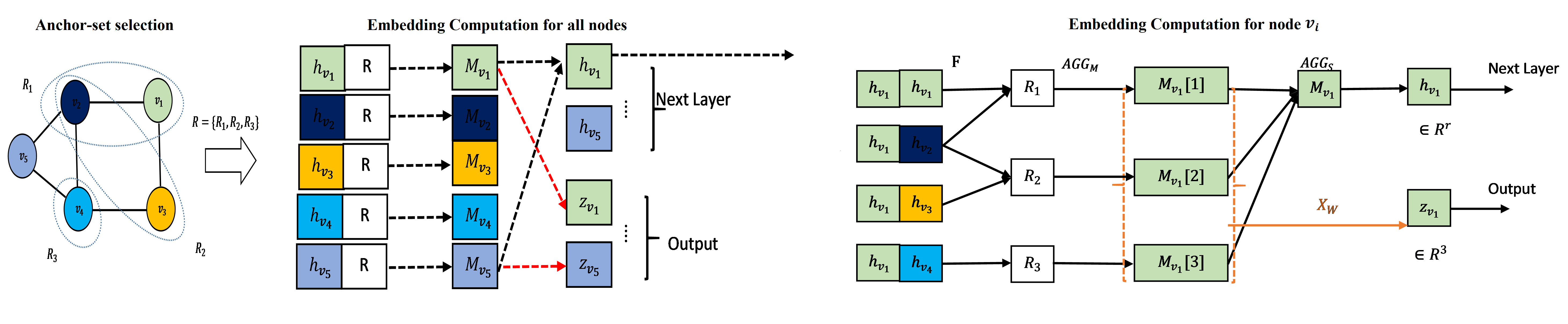}
\caption[P-GNN layer]%
{\small Anchorsets, a subset of graph nodes with different sizes, are randomly selected according to Bourgain’s theorem \cite{bourgain1985lipschitz} $\mathcal{R} = {{R}_1,{R}_2,…,{R}_J}$. In order to compute the embedding for $v_n$, feature information of node $v_n$ and another node from anchor-set ${R}_j$, is fed into message computation function $\mathscr{F}$ with output aggregated through $\text{AGG}_{M}$ to produce $\boldsymbol{M}_{v_n}[j]  ,j=1,2,..,J$ corresponding to ${R}_j$. Then, $\boldsymbol{M}_{v_n}[j]$ is further aggregated through $\text{AGG}_{R}$ to output $h_{{v}_n}$ that is passed to next P-GNN layer. The embedding $\mathbf{z}_{v_n}$ at the last layer is generated by applying the nonlinear $\sigma$ to inner product of $\boldsymbol{M}_{v_n}$ and weight vector $\mathbf{w}$.}
\label{fig:fig1111}
\end{figure*}

We use position-aware GNN which instead of aggregating the information from the nearest neighbours, it aggregates the positional and feature information of each node with randomly selected number of nodes called anchor-sets. Then, the computed message is aggregated across the anchor-sets for each node to incorporate global topological information of the graph in node embedding. Fig. \ref{fig:fig1111} illustrates in detail how the node representation is learned through one P-GNN layer.  The goal is to find the best position-aware embedding $\mathbf{z}_{v}^t$ with minimum distortion for a given node $v$ at time layer $t$.

We have made following modifications to P-GNN:

\begin{itemize}
    \item{\emph{Generalization of P-GNN to Time Varying Graphs}: We adopt the input of P-GNN as supracentrality matrix $\mathbb{C}(\omega)$ that representing a temporal multi-layer graph with $\mathrm{N} \times \mathrm{T}$ number of nodes. We compute an embedding for all nodes in all time layers, since $e_{uv}/x_{uv}/w_{uv}$ defined in (\ref{eq1}) can change from $t$ to $t'$. The embedding $\mathbf{z}_v^t$ will then be aggregated from an RNN based representation to estimate missing data.}

    \item{\emph{Modification of Message Computation Function ($F$)}: In ITS, average speed of neighbouring nodes might correlate more or less at different time layers due to a variety of factors i.e., different types of residential zone, special events, accident, etc. Using attention mechanism while computing an anchor-set's message with respect to a given node $v$ can alleviate misinformative messages from the anchor-set to influence node $v$'s embedding. Therefore, we use the attention mechanism to learn the relative weights between the feature vector of $v$ and its nearest neighbor from the anchor-set. The fact at different degree could apply to other application domains as well. As such, we modify our message computation function to incorporate attention mechanism. 
%
\par In P-GNN, we have multi-level of aggregations as demonstrated in Fig. \ref{fig:fig1111}. First, message of each node $v$ is computed with regards to each anchor-set $R_j$ through function $F$. From each anchnor-set, only the nodes with up to two-hop distant to node $v$ is considered for message computation as shown in the following equation. 

\begin{flalign}
\label{eq17}
\mathcal{M}_j &= \{F(u, v,\mathbf{h}_{u}^{k-1}, \mathbf{h}_{v}^{k-1}) | u \in R_j , R_j \subset \mathcal{R}, \mathcal{R} \sim \mathcal{V},&& \nonumber\\ d(v,R_j) &= {\min}_{u\in R_j} d(u,v) \leq 2\},&& \nonumber \\ 
\boldsymbol{M}_v^k[j] &= \textrm{MEAN}(\mathcal{M}_j), 
\end{flalign}

where $\boldsymbol{M}_v^k[j]$ indicates the aggregated message of $v$ corresponding to $R_j$ that is the average of individual outputs of $F(u, v,\mathbf{h}_{u}^{k-1}, \mathbf{h}_{v}^{k-1})$ where $F$ is defined as:

\begin{eqnarray}
\label{eq117}
F(u, v,\mathbf{h}_{u}^{k-1}, \mathbf{h}_{v}^{k-1})  = \frac{1}{d_{sp}^{q}(u,v)+1}\mathbf{a}_v, 
\end{eqnarray}

where $d_{sp}^{q}(u,v)$ is defined as below:

\begin{eqnarray}
d_{sp}^{q}(u,v) = \left\{\begin{array}{ll}
                d_{sp}(u,v) & \textnormal{if} \ \ d_{sp}(u,v)\leq q, \\
                \infty, & \textnormal{otherwise},\end{array}\right.  
\end{eqnarray}

where $d_{sp}(u,v)$ is the shortest path between a pair of nodes $(u,v)$, and we write $\mathbf{a}_v = w_c(\text{CONCAT}(\mathbf{c}_v,\mathbf{h}_v))$, where $\mathbf{c}_v$ is denoted as
$\mathbf{c}_v = \sum_{u'} \mathbf{\alpha}_{vu'}\mathbf{h}_{u'}$ and $\mathbf{\alpha}_{vu'}$ is computed as following:

\begin{eqnarray}
\label{eq16}
\mathbf{\alpha}_{vu'} & = &\frac{\exp{(\text{score}(\mathbf{h}_{v}^{k-1},\mathbf{h}_{u'}^{k-1})})}{\sum_{u'}\exp{(\text{score}(\mathbf{h}_{v}^{k-1},\mathbf{h}_{u'}^{k-1})})}, 
\end{eqnarray}

where the $\text{score}$ is calculated as:
\begin{flalign}
{\text{score}}(\mathbf{h}_{v}^{k-1},\mathbf{h}_{u'}^{k-1}) &=  V^T(tanh(W_1\mathbf{h}_v^{k-1}+W_2\mathbf{h}_{u'}^{k-1})), \nonumber 
\end{flalign}
where $V$ ,$W_1$, $W_2$ and $w_c$ are trainable weights. $\mathbf{c}_v$ and $\mathbf{a}_v$ are the context vector and attention coefficient, respectively. The is inspired by \cite{bahdanau2014neural}.}

%

\item{\emph{Modification of $AGG_R$}: In P-GNN, $\boldsymbol{M}_{v}[j]$ associated with anchor-set $j$, is averaged across the anchor-sets to generate $\mathbf{h}_{v}$. We choose to differentiate nodes based on their conditional centrality in stationary status $(CC_{v}(\omega)|_{\omega\rightarrow\infty})$, as higher conditional eigenvector centrality indicates higher influence of a given node $v$ and its surrounding nodes compared to the ones with lower eigenvector centrality.  Additionally, corresponding informative anchor-sets contain at least 2-hop neighbour(s) of node $v$, wherein the ones with higher $CC_{v}(\omega)$ deserve to have higher weights for aggregation. As such, 
we substitute $AGG_{R}$ by weighted mean of $\boldsymbol{M}_{v}[j]$, where the weights are proportionate to stationary $CC_{v}(\omega)$ as following:
\begin{eqnarray}
\label{eq18}
\mathbf{h}_v^{(k)}=\frac{1}{J}\sum_{j}r_j \boldsymbol{M}_v[j],\forall j\in[1,J] 
\end{eqnarray}
where $r_j$ is calculated as:
\begin{eqnarray}
r_j &=&\frac{\sum_{u' \in R_j} CC_{{u'}_j}(\omega)}{\sum_{{j'}=1}^J\sum_{u\in R_{j'}}CC_{u_{j'}}(\omega)}|_{\omega\rightarrow\infty}, \nonumber \\ 
&&  d(v,u') \leq 2,
\end{eqnarray}
where $\mathrm{J}$ is the number of anchor sets. Calculation of  conditional node centrality in stationary condition is implemented separately from the main algorithm, and the result has been used in $AGG_R$ to aggregate the computed messages across the anchor-sets. \par Anchor-sets are selected randomly based on Bourgains’ theorem, come in different sizes that distribute exponentially. $\boldsymbol{M}$ is computed for node $v$ from anchor-set $R_j$, if $R_j$ hits $v$, that means $d(v,R_j) \leq 1$. Large anchor-sets have higher probability of hitting $v$, but are less informative of positional information of the node, as $v$  hits at least one of many nodes in the anchor-sets. On the other hand, small anchor-sets have fewer chance of hitting $v$, however, provide positional information with high certainty \cite{you2019position}.}
\end{itemize}

\par At the end of last P-GNN layer, the node embedding $\mathbf{z}_{v}$ is calculated after applying a non-linear $\sigma$ to an inner product of a weight vector $\mathbf{w}$ and $\boldsymbol{M}_{v}$ as in $\mathbf{z}_{v}=\sigma(\mathbf{w}\boldsymbol{M}_{v})$

The output embedding of for all nodes at all layers $t$ can be reformed into a supraembedding matrix $\mathbb{Z}$  as following:.

\begin{equation}
\label{eq20}
\mathbb{Z} =
\begin{bmatrix}
z_{v_1}^1 & z_{v_1}^2 & z_{v_1}^3 & \cdots & z_{v_1}^T\\
z_{v_2}^1 & z_{v_2}^2 & z_{v_2}^3 & \cdots & z_{v_2}^T\\
z_{v_3}^1 & z_{v_3}^2 & z_{v_3}^3 & \cdots & z_{v_3}^T\\
\cdots & \cdots & \cdots & \cdots &\cdots \\
z_{v_{|\mathcal{V}|}}^1 & z_{v_{|\mathcal{V}|}}^2 & z_{v_{|\mathcal{V}|}}^3 & \cdots & z_{v_{|\mathcal{V}|}}^T\\
\end{bmatrix},
\end{equation}
where $\mathbf{z}_{v}^t$ is the embedding vector of node $v$ at time $t$. In some scenarios, the edge embedding is needed rather than node embedding. We estimate $e_(uv)$ embedding by averaging the embedding of the ending nodes as:

\begin{equation}
\label{eq21}
{\mathbf{z}}_{{uv}}=\frac{1}{2}({\mathbf{z}}_{{u}}+ {\mathbf{z}}_{{v}})
\end{equation}
$$$$
We utilize the result of TMP-GNN embedding for missing data estimation.  Before proceeding to our proposed architecture, we briefly review the other components of our framework for missing data. 

\subsection{Bi-directional Recurrent Neural Network (Bi-GRU)} \label{GRU}

\begin{figure}[!t]
\centering
\includegraphics[width=0.55\textwidth]{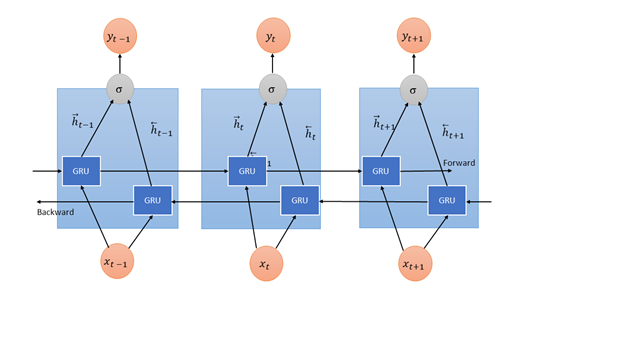}
\caption{Bi-directional Gated Recurrent Unit}
\label{fig:fig111}
\end{figure}

We use Bi-directional GRU as a component in our proposed architecture to estimate missing points in temporal multi-layer graph $\mathcal{G}$.  The Bi-GRU \cite{schuster1997bidirectional}   (Fig. \ref{fig:fig111}) captures the temporal correlation within a time layer in both forward and backward direction. The input to Bi-GRU is a triplet arrays $\boldsymbol{Y}_e$, $\boldsymbol{M}_e$ and $\boldsymbol{\Delta}_e$ that is produced from the graph feature set,  $\mathbb{X} = \{x_{11}^1,x_{11}^2,…,x_{uv}^t,…,x_{uv}^T,…,x_{u'v'}^T \}$, where $x_{uv}^t$ is the feature associated to ${e_{uv}}$ at time layer $t$. We sometimes refer to $x_{uv}$ as $x_e$  interchangeably which $e$ could be any edge in the graph. We also assume that the graph nodes are constant. However, the number of edges and their corresponding features may change from one layer to another. We set $x_{uv}^t$ to 0, if ${e_
{uv}}$ does not exist at time $t$ (${\rho}^{(e_{uv} \times t)} = 0$).  We also select to randomly remove missing points and set it to 0 in $\boldsymbol{Y}_e$. $\boldsymbol{M}_e$ called the mask array containing 0s and 1s that indicates the coordinates of missing and observed points, respectively corresponding to $\boldsymbol{Y}_e$. Additionally, each element in $\boldsymbol{\Delta}_e$ illustrates the time difference between the current and the last layer which the measurement is recorded. $\boldsymbol{\Delta}_e$ is defined to handle the different sampling rate associated with data heterogeneity from different sources \cite{yoon2018estimating} \cite{najafi2020estimation}. Each edge contains $\mathrm{D}$ streams of features. We use $x_{e_d}$ to represent stream $d$ of feature associated to $e$, where $e$ is an edge connecting a pair of nodes in the graph $\mathcal{G}$.
Our goal is to find the best estimate $\hat{x}_{e_d}^t$ with minimum $\text{RMSE}$ for a particular missing point through solving the following optimization problem and finding the function $\text{f}$ as defined below.

\begin{eqnarray}
&& \min_{\text{f}} \EX_{\mathcal{F}} \left[ \sum_{t = 1}^{\mathrm{T}} \sum_{d = 1}^{\mathrm{D}} (1 - m_{e_d}^t) \mathcal{L} ({\mathcal{X}}_{e_d}^t , \widehat{\mathcal{X}}_{e_d}^t)  \right],
\end{eqnarray}
where the loss function is defined as $\mathcal{L} (\mathcal{X}_{e_d}^t , \widehat{\mathcal{X}}_{e_d}^t) = (\{x_{e_d}^t - f_{e_d}^t(\mathcal{X} , \mathcal{T}))^2, \mathcal{X} = \{x_{e_d}^t, \forall e, \forall d, \forall t \}, \mathcal{T} =  \{1, 2, .., \mathrm{T}\}$. 
$\EX$, $\text{f}$ and $\mathcal{F}$ indicate expected value, the desired estimator function and unknown probability distribution that the records are sampled from, respectively. ${\mathcal{X}}_{e_d}^t$ shows stream $d$ of all edges' features at time $t$ and $m_{e_d}$ represents one element of the mask array $\boldsymbol{M}_e$ defined above.

\subsection{E-TMP-GNN: Extended TMP-GNN for Missing Data Estimation} \label{TM-PGNN}

\begin{figure*}[!t]
\centering
\includegraphics[scale=0.15]{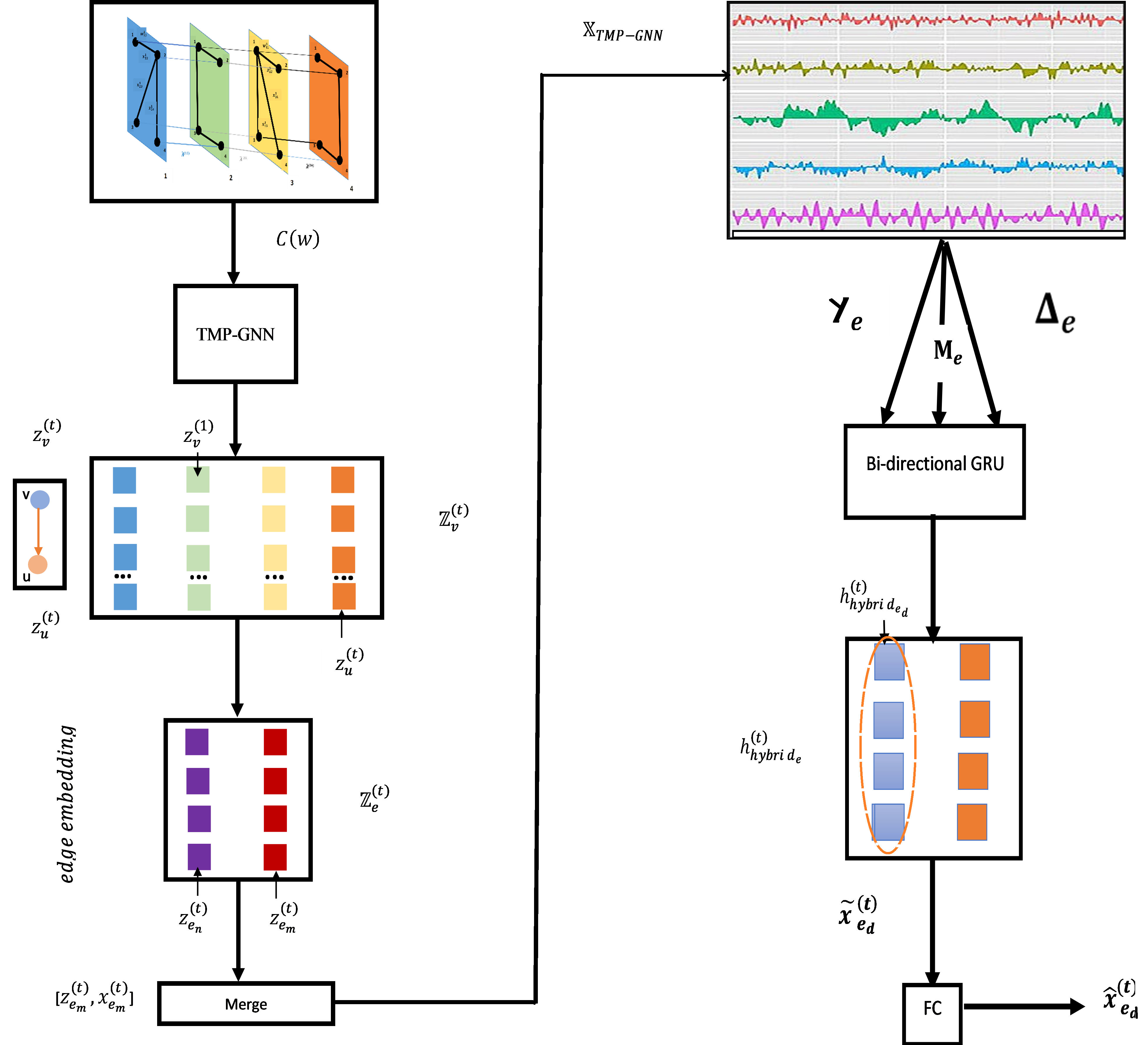}
\caption{E-TMP-GNN I}
\label{fig:fig112}
\end{figure*}

\begin{figure*}[!t]
\centering
\includegraphics[scale=0.12]{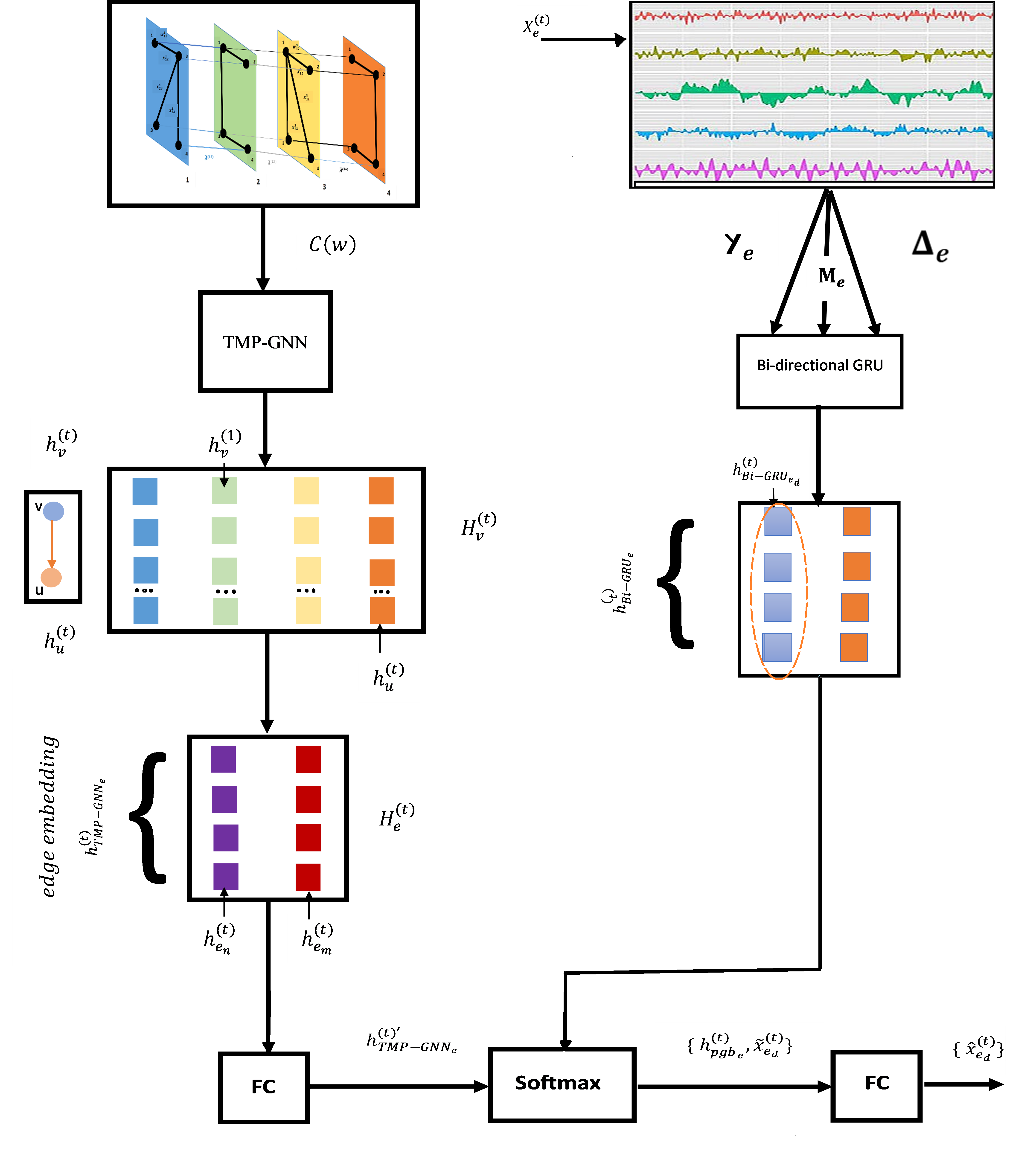}
\caption{E-TMP-GNN II}
\label{fig:fig113}
\end{figure*}

Figs. \ref{fig:fig112} and  \ref{fig:fig113} illustrate our proposed hybrid architecture of TMP-GNN and Bi-GRU. We propose two architectures: E-TMP-GNN I and E-TMP-GNN II. In E-TM-PGNN I, the input is supracentrality matrix $\mathbb{C}(\omega)$ that encodes the intra-layer and inter-layer edge weights of multi-layer graph $\mathcal{G}$. The TMP-GNN's output $\mathbb{Z}$ encodes the topological structure, positional and feature information across the nodes within a layer, and among the same nodes between consecutive layers. Thereafter, the resulted d-dimensional embedding $\mathbb{Z}$ is used to calculate edge embedding as in (\ref{eq20}), then passed to the input of Bi-GRU to merge to existing edge features and form the below array. 
\begin{eqnarray}
{\mathbb{X}}_{\textrm{TMP-PGNN}}=\{x_{e_{{d_i}}}^{(t)}, \mathbf{z}_{e_{{d_i}}}^{(t)} \forall i \in {\mathbb{N}}_{\mathrm{D}}, \forall e \in \mathcal{E}^{(t)}, {\rho}^{(e \times t)} = 1 \}
\end{eqnarray}

\begin{figure}
\centering
\includegraphics[scale=0.45]{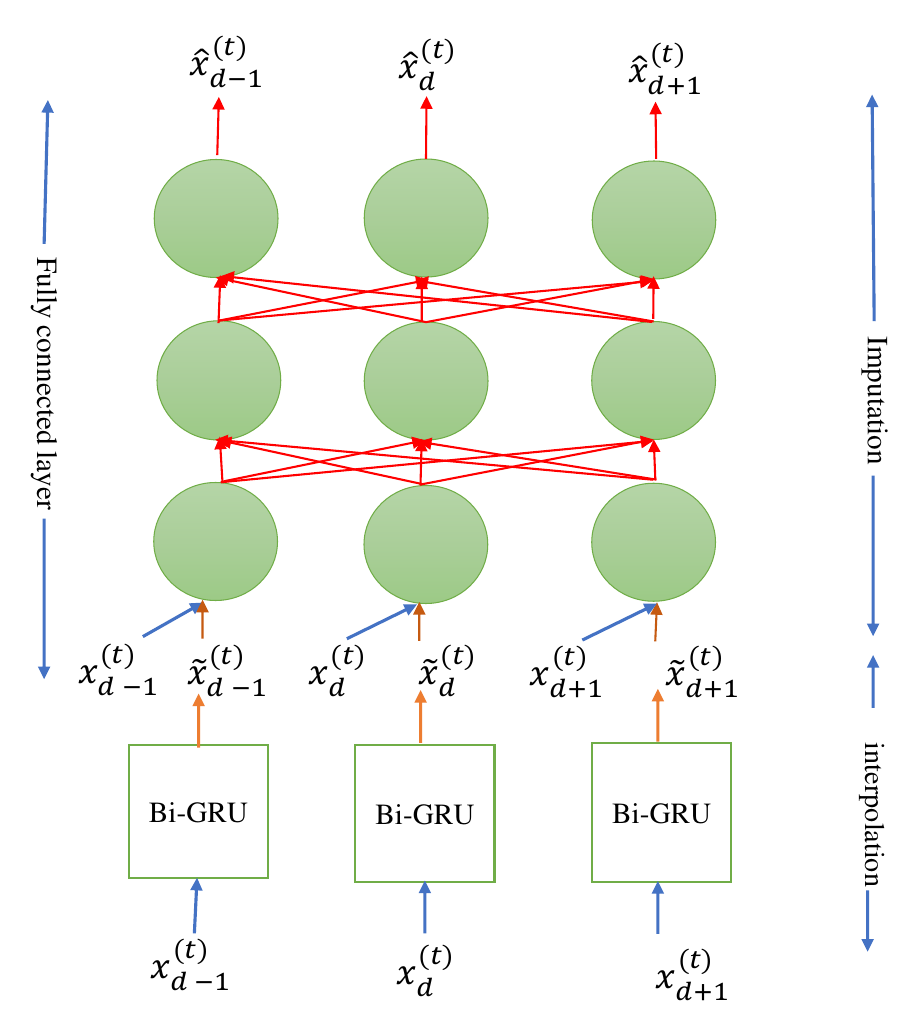}
\caption{M-RNN architecture in time domain \protect\cite{yoon2018estimating}}
\label{fig:fig2}
\end{figure}

 The Bi-GRU is a RNN variant that the output of the previous layer is a part of current layer input. This characteristic let the information propagate step by step \cite{schuster1997bidirectional} and enables the Bi-GRU to capture long-term temporal correlations of edges in two directions which is advantageous to missing data imputation. We use a one layer of Bi-GRU and form a triplet arrays of $\boldsymbol{M}_e$, $\boldsymbol{Y}_e$ and $\boldsymbol{\Delta}_e$ as explained earlier in this section. An initial estimation, $\tilde{x}_{e_d}^t$ is made by applying a nonlinear activation function to the last hidden states of Bi-GRU, $h_{hybrid_{e_d}}^{(t)}$ as below:
 
 \begin{equation}
\label{eq244}
\tilde{x}_{e_d}^t=\sigma(U_d [ \overrightarrow{h}_{hybrid_{e_d}}^{(t)};\overleftarrow{h}_{hybrid_{e_d}}^{(t)}] + g_d),
\end{equation}
where two arrows indicate forward and backward direction of Bi-GRU, respectively. $U_d$ and $g_d$ are weight and bias vectors.
 In order to capture the correlation among sequences of $x_{e_{{d_i}}} |_{i=1}^{\mathrm{D}}$, a fully connected layer with hidden dimension equal to number of feature streams $\mathrm{D}$ is placed afterward. The output of this layer $\hat{x}_{e_d}^t$ is the final estimate of $x_{e_d}^t$. 
 
\begin{equation}
\label{eq245}
\hat{\mathbf{x}}^{(t)} = \sigma(\mathbf{W} \mathbf{h}^{(t)} + \mathbf{\beta_{0}}),
\end{equation}
 
 where $\mathbf{h}_{(t)}$ is defined as following:
 
\begin{equation}
 \label{eq246}
 \mathbf{h}^{(t)} = \phi(U\mathbf{x}^{(t)} + V \mathbf{y}^{(t)} + \gamma_{0}),
\end{equation}
 
 where $\mathbf{y}^{(t)}$ is denoted as $[\mathbf{x}^{(t)}, \mathbf{m}^{(t)}]$. 
 
 The Bi-GRU and FC components apply interpolation and imputation respectively, as they capture the dynamics within a time layer $(\{\mathbf{x}_{e_d}^t, \forall t \in \mathbb{N}_{\mathrm{T}}\})$, and across the streams of features $(\{\{x,\mathbf{z}^{T*}\}_{e_d}^{(t)}, d = \{1, 2, \cdots, \mathrm{D}\}\})$ simultaneously. The combination of Bi-GRU and FC named Multi-directional RNN,inspired by \cite{yoon2018estimating}.
 
\par Our second  proposed architecture is illustrated in Fig. \ref{fig:fig112}. As it seen, the two pipelines have distinct inputs. The input to TMP-GNN is the temporal multi-layer network represented as suppracentrality matrix 
$\mathbb{C}(\omega)$, and we feed Bi-GRU through $\mathbb{X}$. The last $h_{\mathrm{TMP-GNN}_e}^{(t)}$ output from TMP-GNN is then passed through a fully connected layer to reduce the representation dimension as following equation: 
\begin{equation}
\label{eq24}
\mathbf{h'}^{(t)}_{\mathrm{TMP-GNN}_e}=\sigma(\mathbf{w}_p \mathbf{h}^{(t)}_{\mathrm{TMP-GNN}_e}+\mathbf{b}_p),
\end{equation}
where $\mathbf{w}_p$ and $\mathbf{b}_p$ are weight and bias vector, respectively.
The output is then merged by $\mathbf{h}_{\mathrm{Bi-GRU}_e}^{(t)} =[\overrightarrow{\mathbf{h}}_{\mathrm{Bi-GRU}_e}^{(t)};\overleftarrow{\mathbf{h}}_{\mathrm{Bi-GRU}_e}^{(t)}]$ through $\text{softmax}$ attention, $\mathbf{h}_{\mathrm{pgb}_e}$ is generated as following:

\begin{eqnarray}
\label{eq25}
\mathbf{w'}_{p},\mathbf{w'}_b & = & \text{Softmax}(\mathbf{w}_{p},\mathbf{w}_b), \\
\mathbf{h}^{(t)}_{\mathrm{pgb}_e}& = & \mathbf{w'}_{p}\mathbf{h'}^{(t)}_{\mathrm{TMP-GNN}_e}+\mathbf{w'}_{b}\mathbf{h}^{(t)}_{\mathrm{Bi-GRU}_e}
\end{eqnarray}

and an initial estimate of missing value is made as:

\begin{equation}
\label{eq28}
\tilde{x}_{e_d}^{(t)}=\sigma(Q^{d}\mathbf{h}_{\mathrm{pgb}_{e_d}}+\mathbf{\alpha}_0^d)
\end{equation}
A fully connected neural network is used at the end to finalize the prediction of missing values as following:
\begin{equation}
\label{eq26}
\hat{x}_{e_d}^{(t)}=\sigma(\mathbf{w}_{1}\mathbf{h}^{d}_{\mathrm{pgb}_e}+\mathbf{\gamma_{1}})
\end{equation}
The E-TMP-GNN I aims at extracting additional features out of the embedding that yield from TMP-GNN, and use it to further enrich the edge feature sets. We demonstrate in section \ref{result} that the added streams have made significant improvement in $\text{MAE}$ compared to the baseline. TMP-GNN II, however, aims at reducing the number of feature streams by implementing a $\text{softmax}$ layer that gets $\mathbf{h}_{\mathrm{Bi-GRU}_e}^{(t)}$ and $\mathbf{h}_{\mathrm{TMP-GNN}_e}^{(t)}$ as inputs, since $\mathbf{h}_{\mathrm{TMP-GNN}_e}^{(t)}$ is expected to contain valuable information of the temporal correlation among same nodes of consecutive time layers. 

%

\section{Performance Evaluation} \label{result}

We divide this section to three folds. First we review our real world datasets and their associated characteristics in depth. Then, we discuss two potential inputs to our proposed TMP-GNN pipeline and their impact on node embedding performance. Thereafter, we evaluate the performance of E-TMP-GNN I and E-TMP-GNN II on missing data estimation.

\subsection{Datasets} \label{data}

\par We run our experiment on four datasets: TomTom, Covid-19 Mobility (Mobility) \cite{kang2020multiscale}, PhD Exchange (PhD) and Seattle Inductive Loop Detector (Loop Detector). 
 \par $\textbf{TomTom Dataset}$: This dataset contains space mean speed and average travel time during peak congestion for several hundreds of approximately one km long, road segments across the Greater Toronto Area (GTA). The measurements are collected on a minute basis during congestion, where the average space mean speed is less than $70\%$ of vehicular speed during free-flow conditions. We select a three-hour interval from 4:30pm to 7:30pm on Thursday, 8 September 2016 which is expected to have a high proportion of collected records. During this interval, we select road segments with a significant number of mutual timestamps where measurements for a sufficient number of the segments are conducted. From the point of view of multi-layer network, we select 10 layers of GTA road network of 1867 nodes and 985 edges, where each node and edge is a start/ending point of a road segment, and a road segment respectively. We define each edge feature as the ratio of $\frac{\text{space mean speed}}{\text{free-flow speed}}$ and $\frac{\text{free-flow travel time}}{\text{travel time}}$ that is in range of [0, 1]. To reduce the correlation between the consecutive records and better asses the performance of pair-wise node classification in TMP-GNN and missing data estimation, the records are downsampled by rate of $\frac{8}{1}$ ($\delta=\frac{1}{8}$) . The edge weights are also normalized so that the features are comparable during the learning process. Also, the weighted adjacency matrix is created using the highly accurate geolocations of the road segments in the GTA. Once the node embedding is calculated through TMP-GNN, the edge embedding is estimated by averaging the embedding of the corresponding nodes which the edge is connected to as in \ref{eq21}. The resulting embedding will then be used in missing data estimation.

\par $\textbf{PhD Dataset}$: This case of study utilizes a network that represents the exchange of PhD graduates between universities in the fields of mathematical science.  The features we use in our experiments include graduation year of the students who obtained doctoral degree, his/her official academic advisor(s) and the degree-granting university. The aim of former studies is to estimate the flow of doctorate exchanges between universities, individuals who graduate from university $\mathrm{a}$ and then hired at university $\mathrm{b}$. We consider the years $1950–2010$, which result in $\mathrm{T} = 61$ time layers, and there exist a set of $\mathrm{N} = 231$ connected universities during this time span in the united states. In order to build the graph, directed intra-layer edge are created to represent a doctoral degree obtained by a graduate from university $\mathrm{a}$ at year $t$, who later hired as a professor at university $\mathrm{b}$, and at least advised one student there.  In our case, the edge weight indicates the number of graduate doctorates from university $\mathrm{a}$ in year $t$ who later hired at university $\mathrm{b}$ as a faculty member. The edge direction opposes the flow of PhD graduates moving between universities $(\mathrm{b} \rightarrow \mathrm{a})$ \cite{burris2004academic} \cite{taylor2017eigenvector}.
\par $\textbf{Loop Detector Dataset}$ : Our third case of study is the data collected by inductive loop detectors on freeways in Seattle area \cite{cui2018deep} \cite{cui2019traffic}. The dataset contains temporal sequence of time mean speed of the freeway system. The speed information at a desired reference point (milepost) is averaged from multiple loop detectors on the main lanes of the same direction at the specific reference point. Measurements are recorded every 5-minute.  The adjacency matrix calculated in \cite{cui2018deep} is used to convert the data into an undirected graph which edge indicates that a pair of reference points are connected without specifying direction of connectivity. We have downsampled the dataset by rate of $\frac{2}{1}$ ($\delta = \frac{1}{2}$) and selected 11 time layers of the graph.
\par $\textbf{Mobility Dataset}$: Our last case of study is human mobility flow dataset in the U.S. during the COVID-19 epidemic \cite{kang2020multiscale}. To build the graph, directed intra-layer edge is constructed to represent the flow of people between pairs of states.  Each edge indicates the ratio of visits to the population count among the states. We focused on January 1st-22nd 2020 that led to 22 time layers. 

\maketitle
\begin{table*}[!th]

\small
\centering
\captionsetup{justification=centering}
\caption{Characteristics of Datasets}
 \label{tab:ParamValues}
\begin{tabular}{|m{5.35cm}| m{1cm} | m{1cm}|m{1cm}| m{1cm}|}
\hline
 Characteristics & TomTom& Loop Detector & PhD & Mobility\\
\hline 
 $\mathcal{|V|}$  &1867 &323 & 231& 72\\
 $\text{static}:|\mathcal{E}|, \text{dynamic}^*:\sum_{t=1}^{\mathrm{T}}\mathcal{E}^{(t)}$  & 985&1001 &10365* & 2692\\
 $\mathcal{|V|} \times \mathrm{T}$ &18670 & 3553&14091 &1584 \\
 Largest Connected Component (\text{LCC}) &50 & 323& 13847& 1144\\
 No. of Isolated Nodes & 18620 & 3230 & 244 & 440 \\
\hline

\end{tabular}
\end{table*}

\par
\begin{figure}
     \centering
     \begin{subfigure}[b]{0.45\textwidth}
         \centering
         \includegraphics[width=\textwidth]{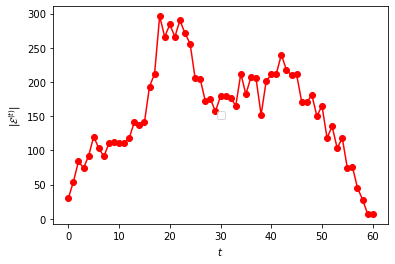}
         \caption{PhD Dataset}
         \label{fig:figa}
     \end{subfigure}
     \hfill
     \begin{subfigure}[b]{0.45\textwidth}
         \centering
         \includegraphics[width=\textwidth]{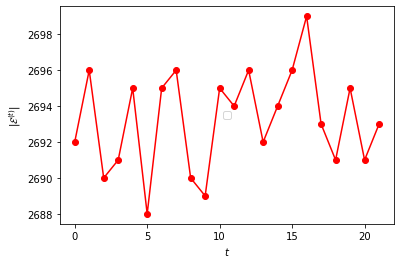}
         \caption{Mobility Dataset}
         \label{fig:figc}
     \end{subfigure}
        \caption{Number of edges per time layer}
        \label{fig:figac}
\end{figure}

 Table \ref{tab:ParamValues} demonstrates the characteristics associated to the graphs built from the 4 selected datasets. 
Among all, TomTom and Mobility have the highest and the lowest number of nodes, respectively. From an edge density's perspective, we have ordered the graph from the most connected to the least connected as following: Mobility, Loop Detector, PhD and TomTom. From the topological point of view, TomTom and Loop Detector are static graph. That is, $\mathcal{E}^{(t)} = \mathcal{E}^{(t')}, t \neq t'$ and the change is in the value of $x_{uv}^{(t)}/w^{(t)}_v$ only. However, PhD and Mobility graphs change throughout the time layers in terms of ${|\mathcal{V}^{(t)}|}, x_{uv}^{(t)}$ and $w^{(t)}_v$. Among the two, PhD has the highest dynamic rate, and turns into the sparsest graph in multiple time layers. Fig. \ref{fig:figac} demonstrates $|\mathcal{E}^{(t)}|$ per layer $t$ for these graphs. 
The third row in the table shows the number of nodes in the temporal multi-layered graph, explained in\ref{supra}, which is equal to $(|\mathcal{V}| \times \mathrm{T})$. TomTom and PhD have significantly largest number of nodes in the temporal graph given the number of available time layers. The fourth row indicates size of largest connected component ($\text{LCC}$) in the dynamic graph. The parameter is considered as an important topological invariant of the graph and can be computed using a variant of depth-first search for directed and undirected graphs \cite{tarjan1972depth}. Here, graphs can be ordered according to ratio of $\frac{\text{LCC}}{|\mathcal{V}|}$ in a descending order as follows: Loop Detector, PhD, Mobility, and TomTom. The entire Loop detector undirected graph is considered as one connected component; TomTom as a directed graph has multiple weakly connected components with each component as small as $2\%$ of the total $|\mathcal{V}|$. We calculate the number of isolated nodes as the nodes which are not connected to largest component by this equation: $\text{no. of isolated nodes} = \mathrm{N} \times \mathrm{T} - \text{LCC}$ that could be another indicator of nodes with lower node degree and thus, sparser graphs. TomTom has the largest number of isolated nodes, as expected.

\subsection{Potential Inputs to TMP-GNN: Single Graph v.s. Multiple Graphs} \label{input}

\par There are several ways we can feed P-GNN: 
\begin{itemize}
    \item{\emph{Supracentrality Graph $\mathbb{C}(\omega)$}: In this case, we represent the centrality of a temporal multi-layer graph via a supracentrality matrix that is analogous to a single graph with $\mathrm{N} \times \mathrm{T}$ number of nodes. The main advantage of this approach is to illustrate the graph with all weighted intra-layer and inter-layer edges. However, the matrix size might get relatively large in the presence of large number of $\mathrm{T}$. In the next section, we demonstrate that this approach outperforms the one with multiple-graphs input.}
    \item{\emph{Multiple Graphs}: In this approach, a sequence of centrality / adjacency matrices associated to each time layer, $\{\boldsymbol{A}^{(1)},\boldsymbol{A}^{(2)}, \cdots,\boldsymbol{A}^{(\mathrm{T})}\}$, is fed into P-GNN. That is, the multi-layer graph is treated as individual instances of single graph without consideration of inter-layer coupling.}
    \item{The input to P-GNN can be a single graph with the entry of its corresponding adjacency matrix as follows:
\begin{equation}
\label{eq14}
a_{uv} = \left\{
     \begin{array}{@{}l@{\thinspace}l}
        [x_{uv}^{(1)},x_{uv}^{(2)},\cdots,x_{uv}^{(\mathrm{T})}]  &: \textnormal{if} \ \ \textnormal{$u, v$ are connected,}\\
        0 &: \textnormal{otherwise.} \\
     \end{array}
   \right.
\end{equation}
In this approach, we treat the multi-layer graph as a single graph where each element of the adjacency matrix is a time series of features associated to an $e_{uv}$. In order to capture the temporal characteristics of a $v$/$e_{uv}$, a recurrent neural network based is required before applying function $F$ in Fig. \ref{fig:fig1111} that leads to high computational complexity. We have not implemented this method.}
\end{itemize}

\par
\begin{figure}
\centering
\includegraphics[scale=0.45]{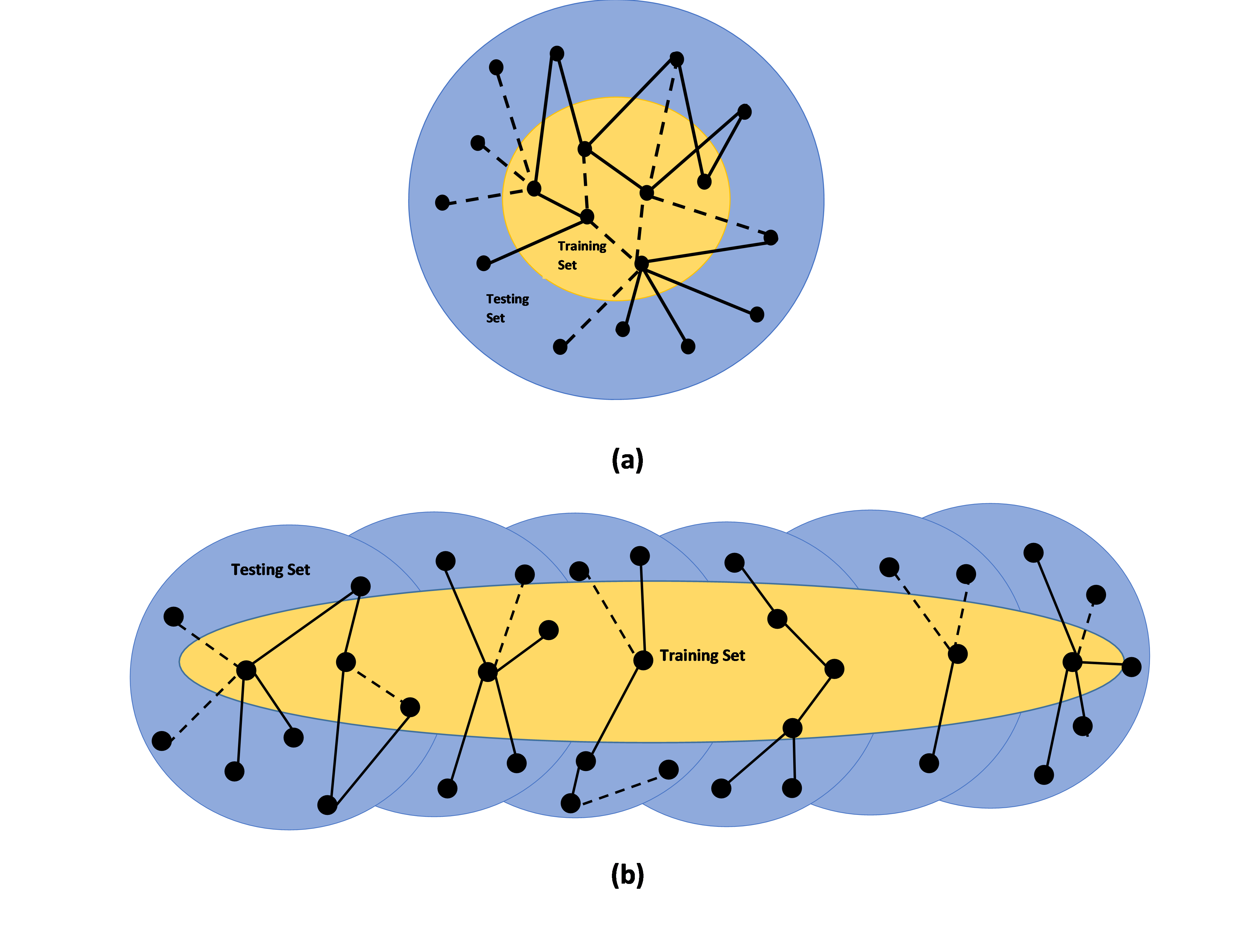}
\caption{train-test split for temporal graph as $\textbf{(a)}$ single graph, $\textbf{(b)}$ a sequence of single layered graph}
\label{fig:figd}
\end{figure}

 $\textbf{Training}$: There are two different ways to train the time-varying graph to compute node embedding. In case of single graph, graph nodes are split into training, evaluation and testing with $80\%$, $10\%$, and $10\%$ respectively.  The classification accuracy on the test set is recorded once the best performance on the evaluation is reached. As for the multiple graphs, there are two techniques to train the graph: first, $80\%$ of the graphs can be used for training and the remaining for evaluation and the test set. Second, $80\%$ of the nodes of all available graphs is used for training, and $10\%$ of the remaining nodes in all time layers of graphs for test and the equal number for validation set. Although both techniques can be utilized for time invariant multiple graphs i.e., Protein \cite{adhikari2020deepcon} \cite{you2019position}. we exploit the second choice for our multi-layer graph venue (Fig. \ref{fig:figd}).  Because, it enables capturing change trajectories throughout the time layers, whereas the first approach split the graphs without considering the time ordering, and also the behavioural changes of any of the nodes in $20\%$ of the graphs is not taken into account. To the best of our knowledge, this is the first time that such experimental setting is used for node embedding in multiple graphs.
\par \textbf{Inductive v.s. Transductive Learning}: There are two different settings to learn the node embedding: Inductive learning and transudative learning. In the latter setting, the graph is trained with fixed node ordering. In this technique, the model needs to retrain, when some additional nodes are added, the ordering has changed or a new graph is given. Also, the node attributes are augmented as one-hot identifiers that could restrict the generalization ability of the model \cite{kipf2016semi}. The authors in \cite{you2019position} asses the transductive learning performance of P-GNN on link prediction task, whether a pair of nodes are connected. On the other hand, in inductive learning, the learned positional information can be transferred to an unseen graph. Particularly, instead of augmenting one-hot identifiers of node attributes, an order invariant scalar could be assigned to nodes, whenever node attributes are not available. We choose to train TMP-GNN in an inductive learning setting and utilize the pair-wise node classification task, as in \cite{you2019position} to demonstrate the performance of TMP-GNN embedding.
\par As discussed in (\ref{pre}), our goal is to compute the embedding for time varying graph with potential change in edge existence ${\rho}^{(\mathcal{E} \times t)}$ or intra-layer edge weights represented in $\boldsymbol{A}^{(t)}$. Therefore, pair-wise node classification would be a more challenging task rather than link prediction, since the node class is determined according to the status of its surrounding edges and/or the status of other isomorphic nodes that is dynamic in such graphs. When the learning task requires node positional information too, only using structure-aware embedding is not sufficient.

\begin{figure}[t!]
     \centering
     \begin{subfigure}[b]{0.45\textwidth}
         \centering
         \includegraphics[width=\textwidth]{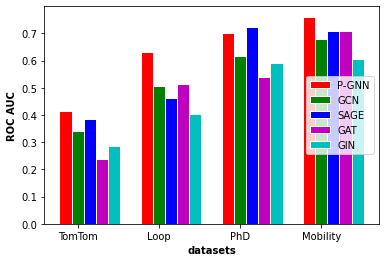}
         \caption{Temporal multigraph input}
         \label{fig:fige}
     \end{subfigure}
     \hfill
     \begin{subfigure}[b]{0.45\textwidth}
         \centering
         \includegraphics[width=\textwidth]{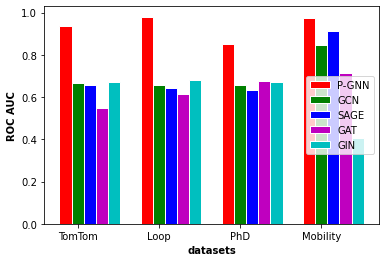}
         \caption{Single supragraph}
         \label{fig:figf}
     \end{subfigure}
        \caption{Pairwise node classification accuracy of TMP-GNN v.s. GNNs}
        \label{fig:figef}
\end{figure}

Fig. 
\ref{fig:figef} illustrates the $\text{ROC AUC}$ of TMP-GNN for pairwise node classification given a second and first type of input, sequences of individual graphs with each graph associated to one layer, and single supragraph, respectively (subsection \ref{input}). All datasets demonstrate a significantly better performance at the presence of a single graph input versus a multigraph input. It is highlighted that single graph has better represented the chronological property of a multi-layer temporal graph. The difference in performance improvement achieved by TMP-GNN is significantly larger in case of single supragraph for the same reason. In case of TomTom, Loop detector and PhD, other four GNN perform relatively the same. As for Mobility, all five GNN-based approaches perform well with relatively negligible difference in classification accuracy. This could be due to lower number of nodes plus lower dynamics in number of edges per time layer $|\mathcal{E}^{(t)}|$ as shown in Fig. \ref{fig:fig1111}, and stronger connectivity within time layer in comparison to other case studies that confine to expressive power of TMP-GNN to less distinct from others. TMP-GNN demonstrates the largest difference of $\text{AUC}$ improvement for TomTom from multigraph to single supragraph. As noted in subsection \ref{data} it has the highest number of nodes with more than 5 times higher than PhD, the second largest graph, in conjunction with lowest connectivity degree, and largest number of isolated nodes which led the discriminating power of TMP-GNN to further underline. 
For three out of four case studies in multigraph implementation (Fig. \ref{fig:fige}), the best performer is TMP-GNN; the second most competent is SAGE. The reason could be the similarity in message computation function $F$, where the feature information of two nodes $(\mathbf{h}_v, \mathbf{h}_u)$ or the attended message from $u$ towards node $v$ are concatenated as following equation:
\begin{eqnarray}
\label{eq:eq30}
&& \hat{\eta}_{\text{y}}^{\text{metric}} = \frac{\left|{\eta}_{\text{y}}^{\text{metric}} - {\eta}_{\text{y'}}^{\text{metric}}\right|}{{\eta}_{\text{y'}}^{\text{metric}}} \times 100 \nonumber,
\end{eqnarray}
where $\eta$ notes the performance, $\text{metric}$ indicates $\text{AUC ROC}$ or $\text{MAE}$ according to the underlying experiment and $|*|$ is absolute value. $\text{y}$ represents TMP-GNN, E-TMP-GNN I or E-TMP-GNN II and $\text{y'}$ is the comparing baseline method.

\begin{table}
    \begin{center} 
    \caption{Performance Comparison of E-TMP-GNN I, E-TMP-GNN II v.s. counterpart methods}
    \label{tab:comparisons}
    \begin{tabular}{ccccc}
    \hline
               &   TomTom       &     PhD   &  Loop &  Mobility \\
    \hline  
     $\hat{\eta}_{\text{multigraph}}^{\text{ROC AUC}}(\%)$      &   7-42  &  0-22  & 18-35 & 6-20 \\
    \hline
    $\hat{\eta}_{\text{Single supragraph}}^{\text{ROC AUC}}(\%)$  & 28-41 & 20-25 & 30-37 & 6-58 \\
    \hline
    $\hat{\eta}_{\text{E-TMP-GNN I}}^{\text{MAE}}(\%)$ & 59-69  & 0-12 & 17-27 & 0  \\
    \hline
    $\hat{\eta}_{\text{E-TMP-GNN II}}^{\text{MAE}}(\%)$ & 94-96 & 29-96 & 0 & 0-15  \\
    \hline
    \end{tabular}
    \end{center} 
\end{table}

\begin{figure}
\centering
\includegraphics[width=0.45\textwidth]{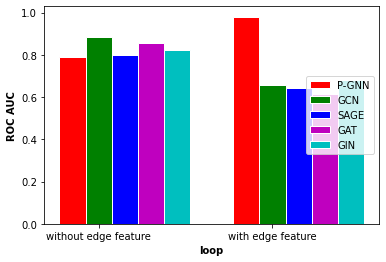}
\caption{Pairwise node classification accuracy of TMP-GNN v.s. GNNs for Loop Detector dataset with and without edge feature}
\label{fig:figg}
\end{figure}

\begin{figure}
     \centering
     \begin{subfigure}[b]{0.44\textwidth}
         \centering
         \includegraphics[width=\textwidth]{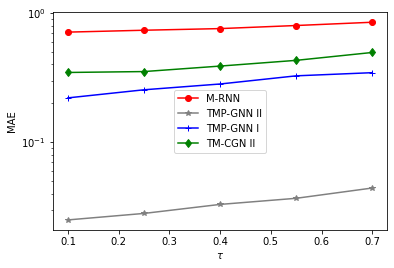}
         \caption{TomTom Dataset}
         \label{fig:figh}
     \end{subfigure}
     \hfill
     \begin{subfigure}[b]{0.44\textwidth}
         \centering
         \includegraphics[width=\textwidth]{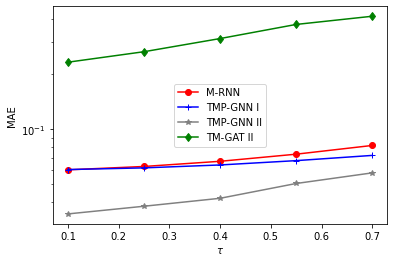}
         \caption{PhD Dataset}
         \label{fig:figi}
     \end{subfigure}
     \hfill
     \begin{subfigure}[b]{0.44\textwidth}
         \centering
         \includegraphics[width=\textwidth]{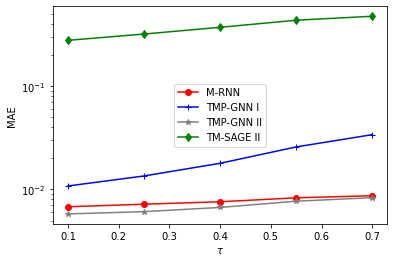}
         \caption{Mobility Dataset}
         \label{fig:figj}
     \end{subfigure}
     \hfill
     \begin{subfigure}[b]{0.44\textwidth}
         \centering
         \includegraphics[width=\textwidth]{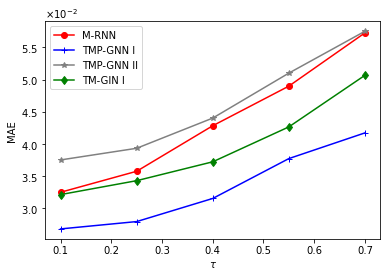}
         \caption{Loop Dataset}
         \label{fig:figk}
     \end{subfigure}
        \caption{Performance Evaluation of E-TMP-GNN I and II on Missing Data Estimation v.s. M-RNN and counterpart GNN}
        \label{fig:fighijk}
\end{figure}

Fig. \ref{fig:figg}  reveals the impact of availability of the edge feature on $\text{AUC ROC}$ for all 5 GNN-based models for Loop Detector dataset. From topological point of view, loop detector is an undirected static graph. When edge features are not available all GNN perform relatively the same.  When taking the time varying edge attributes $x_{uv}^{(t)}$ into account, it turns into a dynamic graph wherein the node class is also dependent on time varying attributes, and TMP-GNN demonstrates its superiority over the other widely used structure-aware embedding. This is where the power of TMP-GNN is highlighted for time varying graphs.
 Table \ref{tab:comparisons} shows the percentage of $\text{AUC ROC}$ improvement made by TMP-PGNN for multigraph and single supragraph implementations as well as the decrease in $\text{MAE}$ made by E-TMP-GNN I and II compared to the baselines for all four datasets where $0$ indicates zero or no improvement.

In Fig. \ref{fig:fighijk}, we evaluate the performance of our proposed architectures, TMP-GNN I and TMP-GNN II on estimation of missing data, and compare it to a RNN variant architecture, named M-RNN (Fig. \ref{fig:fig2}) and the second best classifier demonstrated in Fig. \ref{fig:figf} which is CGN, GAT, GIN and SAGE for TomTom, PhD, Loop and Mobility dataset, respectively. We use Mean Absolute Error (MAE) as a measure of estimation accuracy and select to remove missing values on a temporal dimension. The points are chosen according to uniform distribution in conjunction with a determined missing threshold $\tau$. 

all case studies experience a higher MAE, as $\tau$ increases. 
Moreover, either TMP-GNN I or TMP-GNN II outperforms both M-RNN and the counterpart GNN for each case. This is due to high expressive power of additional features, learning through TMP-GNN representation. The underlying time varying features proved to be significantly more efficient for missing data estimation, than the ones derived from the second-best node classifier in the previous experiment, since it takes position information and interlayer coupling into account simultaneously which the latter makes the architecture more powerful than the ones utilizing RNN variant for capturing long-term temporal dependencies. TomTom has the highest MAE compared to others, where PhD, Loop and mobility come after, respectively. This follows the same order as their corresponding multilayer graphs are ranked according to their degree of connectivity; the sparser the graph is, the larger the estimation error would be. In addition, the percentage of improvement is larger for TomTom, and PhD compared to other two, which means that the proposed method is more effective for graphs with weaker connectivity.

\section{Conclusion} \label{conclusion}
We propose TMP-GNN, a generalized version of P-GNN node embedding for temporal multi-layer graphs that takes short-term temporal correlation as well as feature and positional information into account. We also incorporate the notion of eigenvector based centrality to distinguish nodes with higher influence on their neighbours.  Then, we extended the design to E-TMP-GNN I and II to tackle missing data challenge in dynamic networks with multiple streams of node/edge feature. Conducted experiments on four real world datasets with diverse characteristics demonstrate significant improvements on node-level classification and missing data estimation tasks. It would be interesting to consider formulating the missing data problem directly into an edge representation task for the future work.

\bibliographystyle{IEEEtran}
\bibliography{PgNN}
\vspace{12pt}
\color{red}

\end{document}